\documentclass{article}

\usepackage{microtype}
\usepackage{graphicx}

\usepackage{subcaption}
\usepackage{pifont}
\usepackage{array}
\usepackage{booktabs} 

\usepackage{hyperref}

\usepackage[accepted]{icml2026}

\usepackage{amsmath}

\DeclareMathOperator*{\argmin}{arg\,min}
\usepackage{amssymb}
\usepackage{mathtools}
\usepackage{amsthm}

\usepackage[capitalize,noabbrev]{cleveref}

\theoremstyle{plain}

\theoremstyle{definition}

\theoremstyle{remark}

\icmltitlerunning{\ding{246} ARO: Aligned Representation learning for multi-Omics data}

\usepackage{pifont}

\begin{document}

\twocolumn[
  \icmltitle{\ding{246} ARO: Aligned Representation learning for multi-Omics data}



  \icmlsetsymbol{equal}{*}

  \begin{icmlauthorlist}
    \icmlauthor{Amogh Singh}{equal,cam}
    \icmlauthor{Yash Shah}{equal,cam}
    \icmlauthor{Chiara D'Ercoli}{equal,sap}
    \icmlauthor{Arash Mehrjou}{gsk,max}
    \icmlauthor{Patrick Schwab}{gsk}
    \icmlauthor{Timothy Jones}{cam}
    \icmlauthor{Pietro Liò}{cam}
  \end{icmlauthorlist}

  \icmlaffiliation{cam}{Department of Computer Science, University of Cambridge, Cambridge, UK}
  \icmlaffiliation{sap}{DIAG, University of Rome ``Sapienza", Rome, Italy}
  \icmlaffiliation{gsk}{GSK.ai}
  \icmlaffiliation{max}{Max Planck Institute for Intelligent Systems}

  \icmlcorrespondingauthor{Amogh Singh}{as3600@cl.cam.ac.uk}
  \icmlcorrespondingauthor{Yash Shah}{ys562@cl.cam.ac.uk}
  \icmlcorrespondingauthor{Chiara D'Ercoli}{cd956@cl.cam.ac.uk}
  \icmlcorrespondingauthor{}{dercoli@diag.uniroma1.it}

  \icmlkeywords{Multi-Omic, Machine Learning, ICML}

  \vskip 0.3in
]

\printAffiliationsAndNotice{}  

\begin{abstract}

The high cost of functional molecular assays, and prevalence of missing modalities and unmatched samples in computational biology, create significant barriers to comprehensive multi-omic profiling, essential for capturing and reasoning over molecules, cells, tissues, and organisms. 
This work proposes a model that learns meaningful representations from multi-omics cancer data supporting the reconstruction of missing and unpaired modalities. Contrary to increasingly complex, larger models, e.g.~Foundation Models (FMs), ARO prioritizes practical applicability in limited or incomplete data settings.
ARO optimally reconstructs missing modalities (MSE of $0.15$ on the validation and test data in the Unmasked settings), with its learned latent embeddings enabling a downstream cancer classification task. Our findings indicate that analyzing diverse molecular layers as a single integrated system offers a reliable and cost-efficient approach, reducing dependence on large-scale experimental testing, while still supporting multi-omic exploration in limited data settings.
\end{abstract}

\section{Introduction}

A common bottleneck in computational biology is the limited availability of paired, complete multi-omics datasets necessary to fully characterize biological processes. 
Prohibitively expensive assays bottleneck the acquisition of high-quality multi-omic data, as they require large sample volumes, and are harder to find in multiple paired modalities \cite{MLinraredisease}. Transcriptomic measurements constitute the most accessible data modality; however, they provide only a partial view of cellular function, not fully reflecting downstream biological processes shaped by post-translational modifications \cite{fernie} and metabolic flux. This makes the integration of multiple omic layers essential for comprehensive understanding of disease mechanisms and therapeutic response, also in the context of rare diseases \cite{subramanian}. 

FMs and multi-omics data integration approaches have revolutionized the field by enabling self-supervised learning on several datasets that capture biological principles and derive joint representations of data. State-of-the-art FMs exploit information from omics modalities including imaging data or structural motifs \cite{uhler, mot}. However, by compressing these biologically rich layers into rigid structural abstractions or secondary imaging proxies, such architectures often overlook the dynamic signaling networks and regulatory feedback mechanisms that underpin cellular homeostasis and organisms representation.

Although existing approaches to multi-omics integration, such as Denoising Autoencoders (DAE) and Multi-Kernel Learning (MKL) have proven effective \cite{denoisingmultikernel}, they are typically tailored to narrow, task-specific settings. Moreover, they assume complete, well-aligned multi-omics data \cite{madrigal}, or focus on robustness to missing subsets of modalities, rather than complete modality recovery \cite{modisworkshop}. 

Despite these advances, cross-modal inference, such as inferring proteomics or metabolomics from transcriptomics, is still an open problem. To address these limitations, this work aims to present the following contributions: 
\begin{enumerate}
    \item Introducing a model able to process a combination of transcriptomic, proteomic and metabolomic layers that does not require paired modality samples, and capable of mapping accessible transcriptomic profiles to predicted proteomic and metabolomic spaces.
    \item Showing that ARO learns a shared latent representational space from which it can correctly infer data characteristics.
    \item  Demonstrating that when testing the proposed model on several downstream tasks, ARO achieves 100\% accuracy on classifying tumorous samples and 72.4\% on laterality classification.
\end{enumerate} 

\section{Background and Preliminaries}
\begin{figure*}[h]
  \vskip 0.2in
  \begin{center}
    \centerline{\includegraphics[width=0.8\textwidth]{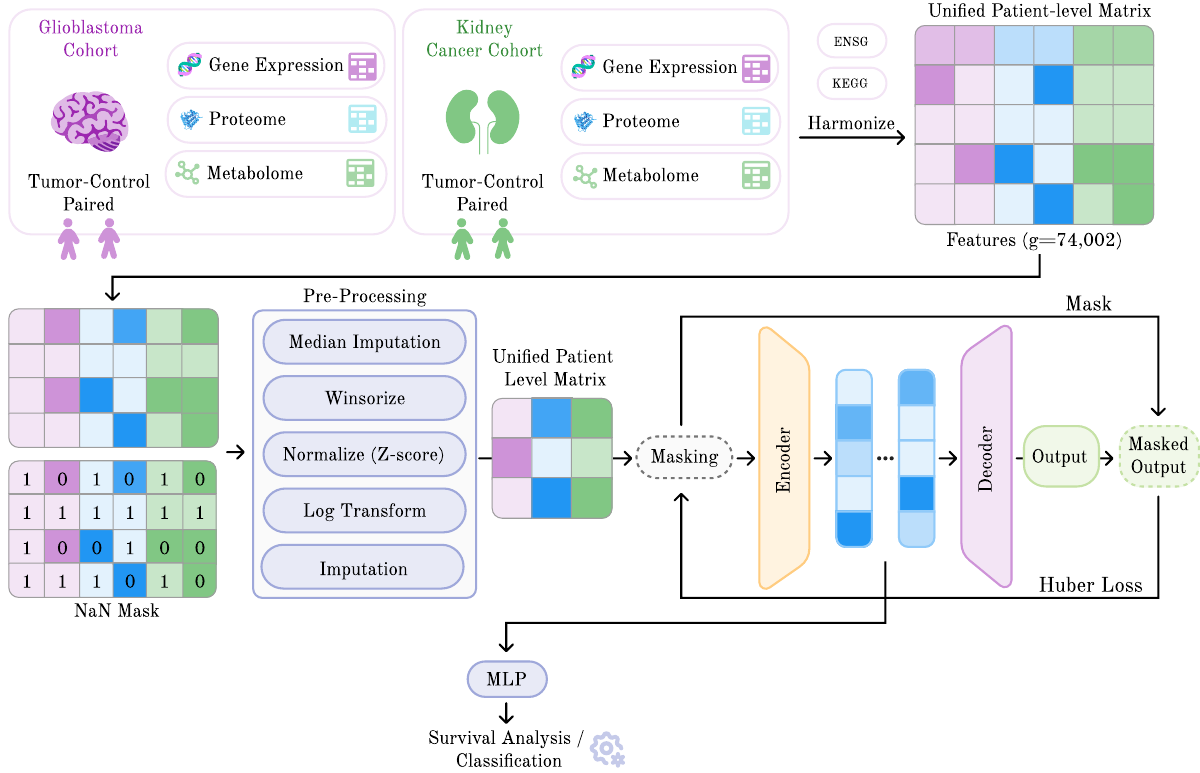}}
    \caption{Overview of ARO. Multi-omics cancer datasets are first harmonized and aggregated. Next, pre-processing involving median imputation, Winsorization, normalization, and log transformation is performed. Features are then optionally masked (shown by dotted outlines) and passed to the model. The loss is computed, and learned latent representations can be used for downstream tasks.}
    \label{fig:architecture}
  \end{center}
\end{figure*}
Current computational modeling research of omics-modalities focuses on developing FMs exploiting the transformer architecture \cite{transformers}, aimed at extracting and interpreting patterns from omics data, rather than reconstructing missing modalities or completing multi-omics profiles. These models are typically restricted to specific data modalities, including uni-modal \cite{dnabert, massgenie, nucleotidetransformer}, single-cell \cite{scgpt, teddy}, protein structure–informed \cite{deepdtf}, or imaging datasets \cite{uhler}.

Recent multi-modal approaches for drug response prediction \cite{deepdtf} and other task-specific applications \cite{mot} have begun integrating heterogeneous data sources through fusion techniques and transformers. However, these methods often combine fundamentally different modalities (e.g.~structural and sequence data) rather than focusing on multi-omics alignment within the same-modality (e.g.~tabular).

To ensure biologically coherent integration of modalities, some approaches leverage domain priors such as knowledge graphs and molecular pathways, while others redirect their focus on cross-species integration \cite{precious3gpt, madrigal}.
Additionally, recent work has leveraged Autoencoders (AEs) specialized to each modality to target omics-representation \cite{modisworkshop}.

\section{Data Extraction and Pre-processing} \label{sec:data_process}
ARO reconstructs missing and unpaired modalities from a single observed input. Specifically, it enables the reconstruction of proteomic and metabolomic data from transcriptomic measurements. Unlike existing approaches, ARO employs a single autoencoder shared across modalities; does not require pretraining on large external datasets, and does not incorporate biological priors, pathway information, or auxiliary imaging data. Instead, through careful pre-processing (modality alignment, handling of missingness, and normalization), a unified latent representation (a single AE), and masking, it enables accurate reconstruction of masked data.

The study leverages two publicly available multi-omics cancer datasets: (1) a glioblastoma cohort \cite{data_glioblast}, including gene expression (FPKM), proteome, phospho-proteome, and metabolome, as well as (2) a renal cell cancer cohort \cite{data_kidney}, including RNA-seq expression (TPM), protein abundance, and metabolite compound abundance. For both cohorts, multi-omics profiles are obtained for tumor and normal paired samples. 

Each modality is represented as a feature matrix, and these matrices are concatenated along the feature dimension to form a unified dataset of size 115 × 74,002, where each feature corresponds to a distinct molecular entity spanning transcriptomic, proteomic, and metabolomic measurements, and each row to a patient. Feature identifiers are harmonized to ENSG gene IDs via g:Profiler \cite{gProfiler}, and KEGG metabolite IDs via MetaboAnalyst with stripped ENSEMBL IDs version suffixes. Features with zero variance (e.g.~all-zero columns obtained after NaN imputation) are removed, therefore showing a different parameter counts for the model in Table \ref{paramstable}. Train-validation-test splits are performed at the patient level considering a 70/15/15 ratio.

Missing value handling and normalization is also performed. NaN positions are recorded into binary masks and filled with per-feature medians from the training-set. Next, winsorization reduces the influence of extreme outliers, by clamping values to the 1st and 99th percentiles (computed on the training data). Respecting the distinct magnitudes of each omic modality, Z-score normalization is utilized to perform per-feature standardization using training-set mean and standard deviation. Log transform (log1p) is applied after clamping to non-negative values. A second round of median imputation addresses NaNs introduced by Scipy's Z-score standardization (stable values for specific gene types, result in low variance features that cause a division by zero). The resulting unified matrix is then used as input for the modality masking step and subsequent model training. 

Regarding normalization, we acknowledge that the two cohorts considered originate from different transcriptomic quantification pipelines (FPKM in \cite{data_glioblast} and TPM in \cite{data_kidney}). Rather than harmonizing the datasets by converting them to a common expression metric, we adopted an approach more aligned with machine learning practice than with traditional bioinformatics workflows. Specifically, normalization was performed independently for each feature, allowing each gene to be standardized according to its own distribution within the respective dataset. Furthermore, the primary objective of this work is methodological rather than bioinformatic. Our aim was not to construct a rigorously harmonized multi-cohort expression dataset, but rather to investigate whether an autoencoder-based framework is able to learn meaningful latent representations from heterogeneous omics data under minimal domain-specific preprocessing assumptions.

We would like to emphasize that care is taken to prevent leakage of any information regarding the validation or test set distributions into the training data.

\section{Architecture}
ARO is a masked autoencoder designed to learn robust representations from high-dimensional multi-omics data, as illustrated in Figure \ref{fig:architecture}. 
It employs a multi-layer perceptron (MLP)-based \cite{multilayerperceptr} encoder/decoder, and a fully-connected hidden layer stack with ReLU non-linearities. Details are provided in Appendix \ref{architecturedetails}.

Two setups are used for training ARO: (1) masking all omics modalities in the input apart from transcriptomics (\begin{sc}ARO Masked\end{sc}), and (2) omitting input masking (\begin{sc}ARO Unmasked\end{sc}). Input masking allows ARO to reconstruct the other two omic-modalities using only transcriptomics.

During training, a random subset of features (10\%) is dropped at each forward pass by setting their values to zero. The model is then optimized to reconstruct the original input via the decoder. 
Reconstruction quality is optimized using the Huber loss \cite{huberloss} (defined in Equation \eqref{eq:huberloss}), with missing values (NaNs) excluded from this loss via a binary mask, ensuring that the model is not penalized for imputing unobserved data. Optimization and hyperparameter tuning details are provided in Appendix \ref{hyperparamtuning}.
Performance is evaluated through Mean Squared Error (MSE), together with downstream classification metrics and confusion matrices. 

\begin{figure*}[h]
    \centering
    \begin{subfigure}{.3\linewidth}
        \includegraphics[width=\linewidth]{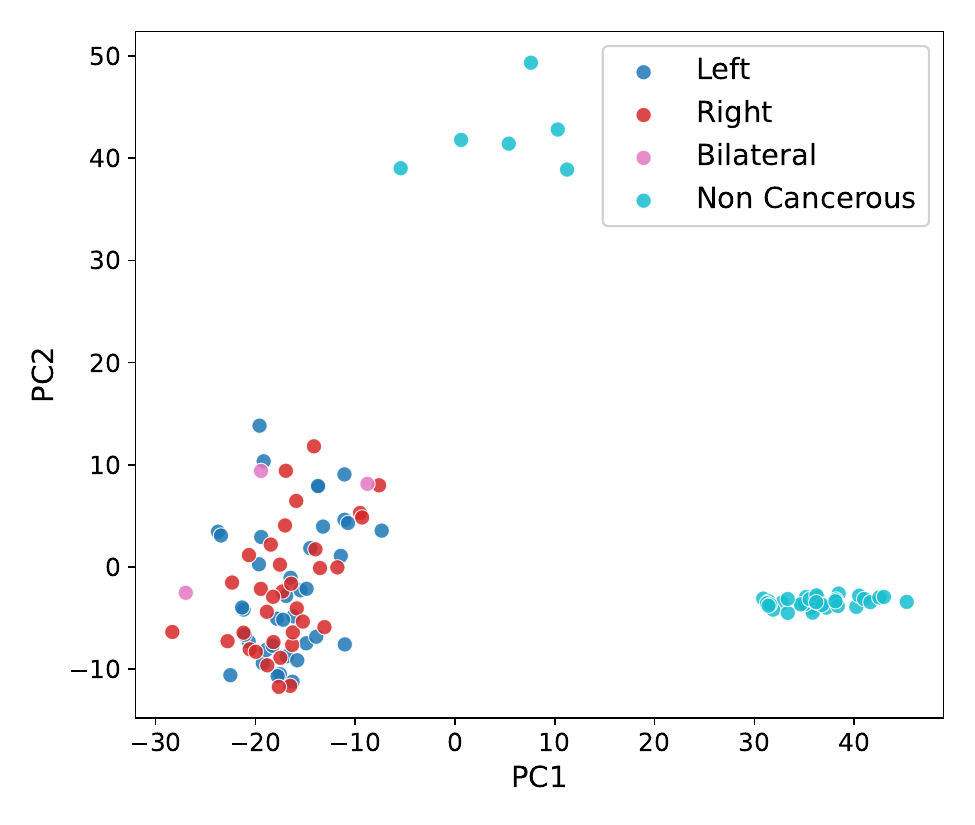}
        \caption{}
        \label{fig:pca_cluster}
    \end{subfigure}
    \begin{subfigure}{.3\linewidth}
        \includegraphics[width=\linewidth]{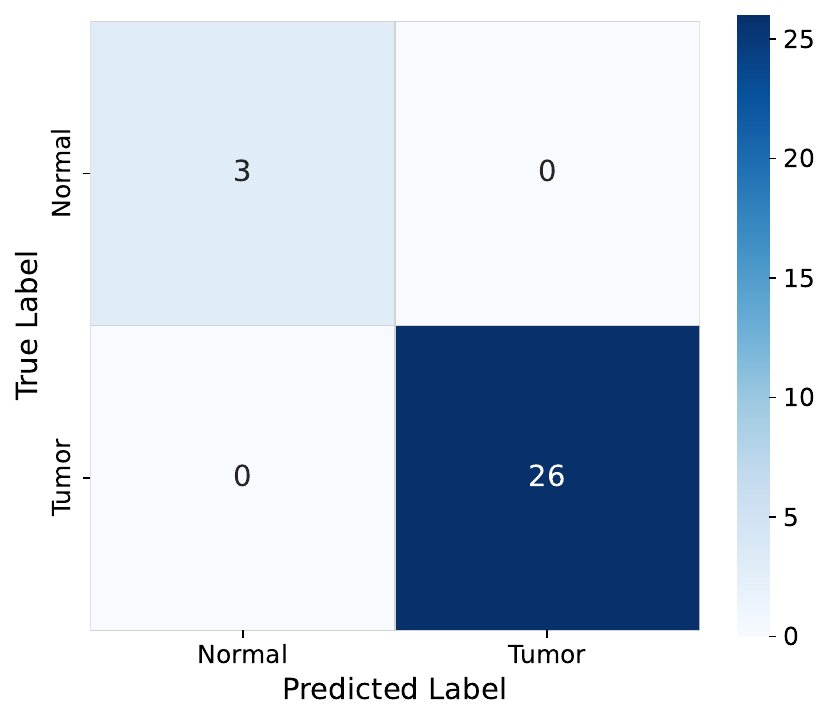}
        \caption{}
        \label{fig:cancer_conf}
    \end{subfigure}
    \begin{subfigure}{.3\linewidth}
        \includegraphics[width=\linewidth]{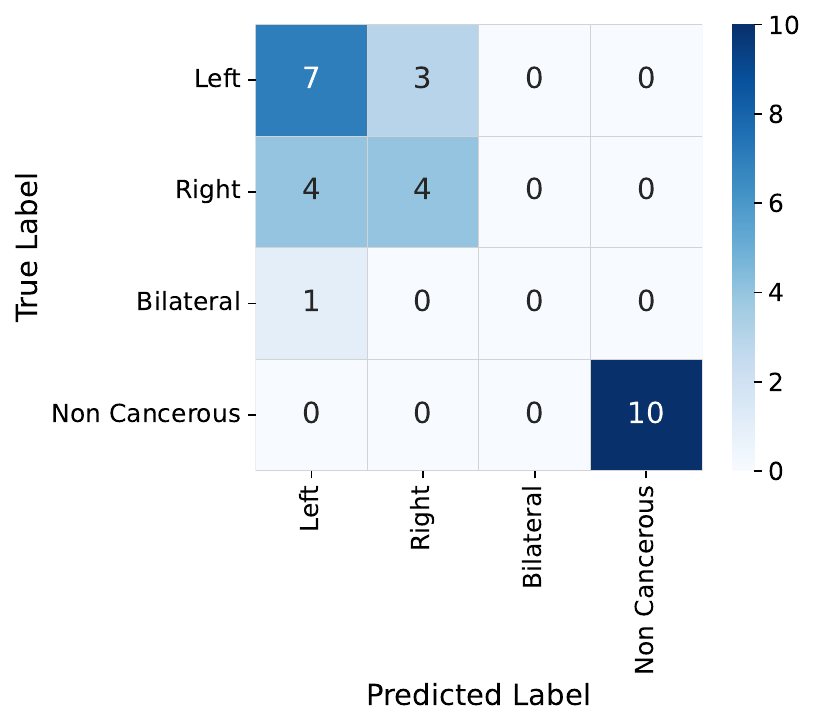}
        \caption{}
        \label{fig:laterality_conf}
    \end{subfigure}
    \label{fig:results_fig}
    \caption{\textbf{ARO performance} in discriminating each omics-modality within the latent space, visualized using PCA (Figure \ref{fig:pca_cluster}) as well as across downstream tasks including binary cancer classification (Figure \ref{fig:cancer_conf}) and cancer laterality prediction (Figure \ref{fig:laterality_conf}).}
\end{figure*}

\section{Experiments and Results} \label{sec:results}
\textbf{Reconstruction}
ARO is designed to reconstruct omic profiles in both masked and unmasked settings, where the masked setting involves inferring the remaining omic modalities solely from transcriptomic input. Given the combination of imputation and data handling during the preprocessing step, we ensured that the model did not exhibit any bias toward trivial solutions--such as predicting median-valued outputs. In particular, since the reconstruction loss is computed only over observed values (non-masked), and the masking pattern varies across samples and batches, the model is discouraged from converging to a trivial statistic. This is empirically demonstrated by the fact that the reconstructed outputs preserve the underlying distributional structure of the data rather than collapsing to mean-like predictions.

Reconstruction performance of masked and unmasked ARO compared to PCA, is presented in Tables \ref{tab:mse_comparison} and \ref{tab:huber_comparison} and the full results are in Tables \ref{tab:appndx_mse_comparison} and \ref{tab:appndx_huber_comparison}. Although PCA achieves superior reconstruction accuracy, its performance on downstream tasks, discussed subsequently, is substantially weaker. 

After identifying the framework proposed in \cite{modisworkshop} as the closest approach to ARO, we aimed at comparing the two methods. However, due to the high dimensionality of our data, the MODIS architecture scales to approximately 22B parameters, resulting in out-of-memory (OOM) errors and rendering complete training infeasible. Constructing a reduced-scale version of MODIS was not a meaningful option. The high dimensionality of our dataset primarily arises from the large number of gene features rather than the number of samples--therefore, reducing the model size would have required discarding a substantial portion of the omics set, fundamentally altering the problem setting and compromising validity of the comparison.  
In contrast, ARO remains computationally tractable across a plausible parameter range ($\approx$50M–700M parameters), with model size showing limited impact on reconstruction performance, thereby enabling more efficient training in high-dimensional regimes. Our limited MODIS training achieved an MSE loss of $2.53$, $0.55$ and $0.43$ for transcriptomics, proteomics, and metabolomics respectively. This further credits the robust performance of ARO compared to existing work.

\begin{table}[h]
  \caption{Reconstruction performance using MSE Loss averaged over 5 runs}
  \label{tab:mse_comparison}
  \centering
      \begin{sc}
        \begin{tabular}{lccc}
          \toprule
             & Train & Validation & Test \\
          \midrule
          ARO Masked  & $0.046$ & $0.14$ & $0.14$ \\
          ARO Unmasked  & $0.064$ & $0.15$ & $0.15$ \\
          PCA  & $3.1e-30$ & 0.14 & 0.14 \\
          \bottomrule
        \end{tabular}
    \end{sc}
\end{table}

\begin{table}[h]
  \caption{Reconstruction performance using Huber Loss averaged over 5 runs}
  \label{tab:huber_comparison}
  \centering
        \begin{sc}
        \begin{tabular}{lccc}
          \toprule
             & Train & Validation & Test \\
          \midrule
          ARO Masked  & $0.022$ & $0.070$ & $0.068$ \\
          ARO Unmasked  & $0.032$ & $0.071$ & $0.07$ \\
          PCA  & $1.5e-30$ & $0.068$ & $0.065$ \\
          \bottomrule
        \end{tabular}
        \end{sc}
  \vskip -0.1in
\end{table}

\textbf{Shared representation}
Embeddings extracted from the latent space of our trained models are used to generate PCA (Figure \ref{fig:pca_cluster}) and t-SNE plots (Figure \ref{fig:tsne_cluster}) that show distinct clusters of cancerous and non-cancerous samples, and of cancer laterality. The clear separation between sample groups indicates that the learned embeddings effectively capture the underlying structure of the original omics data while retaining sufficient discriminatory information to distinguish between clinical states.

\textbf{Downstream analysis}
For evaluating the performance of ARO on downstream tasks, the embeddings from the hidden layer of the \textsc{ARO Unmasked} model are used with an MLP to perform binary classification (cancer/non-cancer) and multi-label classification (cancer laterality). ARO performs well on the binary classification task, achieving 100\% accuracy on test data, and it accurately distinguishes non-cancerous from cancerous samples, as shown in Figures  \ref{fig:cancer_conf} and \ref{fig:laterality_conf} respectively. Although ARO demonstrated ability to infer tumor laterality, with an accuracy of 72.4\%, performance on this task was comparatively weaker---an outcome that aligns with biological expectations, as predicting laterality requires more subtle and complex biological signals (i.e.: spatial or micro-environmental) that cannot be inferred from omics profiles alone.
In comparison, using the projections from PCA for downstream tasks gives 72.41\% and 58.62\% accuracy on the binary and multi-label classification tasks respectively, and the corresponding confusion matrices are shown in Figure \ref{fig:pca_multi_conf_mat} and Figure \ref{fig:pca_binary_conf_mat}.

\section{Limitations and Future Work}

Although the data are highly feature-dimensional, the number of training samples remains limited. This setting may constrain the model’s ability to generalize to unseen data distributions, highlighting the need for future validation on external cohorts, possibly also including other diseases.

ARO's low-capacity architecture enables effective training with limited samples, reflecting the intended scope of the model. The focus of this work is not on architectural innovation, but on robust learning from aligned multi-omics matrices, in contrast to heterogeneous multi-modal settings such as imaging or spatial transcriptomics. The learned associations remain statistical regularities and carry no claims of biological plausibility. Further analysis would be necessary to substantiate potential biological meaning.

Finally, an extension of ARO could include the integration of generative settings, such as variational autoencoders (VAEs), to enable the full synthesis of proteomic and metabolomic profiles conditioned on transcriptomic inputs.

\section{Conclusion}
ARO captures meaningful biological structure across heterogeneous omics modalities and enables reconstruction of missing modalities from a single input. By training ARO to reconstruct masked inputs, robust shared representations are learned.
Notably, even when trained solely on the reconstruction objective, the learned latent representations exhibit well-separated clustering structure using linear projections (PCA) and non-linear techniques (t-SNE). The obtained embeddings enable accurate discrimination between cancerous and non-cancerous samples and qualitative performance on cancer laterality classification using lightweight classifiers such as MLPs.

These results highlight the promise of masked reconstruction as a foundation for multi-omics representation learning, motivating future work to evaluate generalization across datasets and disease types, as well as to explore generative extensions of ARO, such as Large Language Models (LLMs) through the obtained representations.

The potential of this work lies in learning a unified representation across multiple omic modalities within a single latent space, instead of combining separate representations from each modality after training. Given the increasing availability of open biomedical datasets and the growing interest in computational biology modeling, we believe that these developments may contribute to more comprehensive solutions for paired multi-omics data collection.

\section*{Software and Data}

This study leverages publicly available multi-omics data from two distinct cancer cohorts: (1) glioblastoma \cite{data_glioblast} and (2) renal cell carcinoma \cite{data_kidney}, to evaluate the model's capability for cross-omic translation and latent representation alignment. The complete source code is available at \href{https://github.com/ccdderc/ARO}{https://github.com/ccdderc/ARO}.

\section*{Impact Statement}

The broader impact of this work lies at the interface of machine learning and computational biology. By enabling the reconstruction of entire omics modalities---such as proteomics and metabolomics---from more readily available transcriptomic data, the proposed approach addresses the limitation of paired dataset scarcity in multi-omics research and offers the potential to expand the utility of existing data resources. 
However, reconstructed modalities should be interpreted as approximations rather than direct biological measurements. Care must be taken to validate results in clinical or high-stakes applications, to avoid over-reliance on inferred data. 

\section*{Contributions}
\textbf{Amogh Singh} contributed to the multi-omic data selection process, development of the model, collection of the comparison metrics and the analysis to validate the performance on downstream tasks.
\textbf{Yash Shah} also contributed to the implementation of the primary autoencoder approach, including the data pre-processing pipeline, and explored alternative approaches, namely the use of variational autoencoders and transformers.
\textbf{Chiara D'Ercoli} also contributed to the multi-omic data selection process, including data acquisition and preparation and efforts to source additional data, literature review, drafting, writing, and finalizing the manuscript.    

\bibliography{main_paper}

@article{fernie,
  title={On the discordance of metabolomics with proteomics and transcriptomics: coping with increasing complexity in logic, chemistry, and network interactions scientific correspondence},
  author={Fernie, Alisdair R and Stitt, Mark},
  journal={Plant Physiology},
  volume={158},
  number={3},
  pages={1139--1145},
  year={2012},
  publisher={American Society of Plant Biologists}
}

@article{teddy,
  title={TEDDY: a family of foundation models for understanding single cell biology},
  author={Chevalier, Alexis and Ghosh, Soumya and Awasthi, Urvi and Watkins, James and Bieniewska, Julia and Mitrea, Nichita and Kotova, Olga and Shkura, Kirill and Noble, Andrew and Steinbaugh, Michael and others},
  journal={arXiv preprint arXiv:2503.03485},
  year={2025}
}

@article{modisworkshop,
  title={MODIS: multi-omics data integration for small and unpaired datasets},
  author={Lepe-Soltero, Daniel and Arti{\`e}res, Thierry and Baudot, Ana{\"\i}s and Villoutreix, Paul},
  journal={arXiv preprint arXiv:2503.18856},
  year={2025}
}

@article{subramanian,
  title={Multi-omics data integration, interpretation, and its application},
  author={Subramanian, Indhupriya and Verma, Srikant and Kumar, Shiva and Jere, Abhay and Anamika, Krishanpal},
  journal={Bioinformatics and biology insights},
  volume={14},
  pages={1177932219899051},
  year={2020},
  publisher={SAGE Publications Sage UK: London, England}
}

@incollection{huberloss,
  title={Robust estimation of a location parameter},
  author={Huber, Peter J},
  booktitle={Breakthroughs in statistics: Methodology and distribution},
  pages={492--518},
  year={1992},
  publisher={Springer}
}

@article{MLinraredisease,
  title={Machine learning in rare disease},
  author={Banerjee, Jineta and Taroni, Jaclyn N and Allaway, Robert J and Prasad, Deepashree Venkatesh and Guinney, Justin and Greene, Casey},
  journal={Nature methods},
  volume={20},
  number={6},
  pages={803--814},
  year={2023},
  publisher={Nature Publishing Group US New York}
}

@article{denoisingmultikernel,
  title={Multi-Omics Data Integration for Improved Cancer Subtyping via Denoising Autoencoder-Based Multi-Kernel Learning},
  author={Yao, Xiukun and Wang, Tong and Yang, Qi and Wang, Jiawen and Qi, Yao and Xu, Tong and Wei, Zhiwen and Cui, Yuehua and Cao, Hongyan and Yun, Keming},
  journal={Genes},
  volume={16},
  number={11},
  pages={1246},
  year={2025},
  publisher={MDPI}
}

@article{multilayerperceptr,
  title={Learning representations by back-propagating errors},
  author={Rumelhart, David E and Hinton, Geoffrey E and Williams, Ronald J},
  journal={nature},
  volume={323},
  number={6088},
  pages={533--536},
  year={1986},
  publisher={Nature Publishing Group UK London}
}

@article{autoencoders,
  title={Auto-association by multilayer perceptrons and singular value decomposition},
  author={Bourlard, Herv{\'e} and Kamp, Yves},
  journal={Biological cybernetics},
  volume={59},
  number={4},
  pages={291--294},
  year={1988},
  publisher={Springer}
}

@article{transformers,
  title={Attention is all you need},
  author={Vaswani, Ashish and Shazeer, Noam and Parmar, Niki and Uszkoreit, Jakob and Jones, Llion and Gomez, Aidan N and Kaiser, {\L}ukasz and Polosukhin, Illia},
  journal={Advances in neural information processing systems},
  volume={30},
  year={2017}
}

@article{uhler,
  title={Multi-domain translation between single-cell imaging and sequencing data using autoencoders},
  author={Yang, Karren Dai and Belyaeva, Anastasiya and Venkatachalapathy, Saradha and Damodaran, Karthik and Katcoff, Abigail and Radhakrishnan, Adityanarayanan and Shivashankar, GV and Uhler, Caroline},
  journal={Nature communications},
  volume={12},
  number={1},
  pages={31},
  year={2021},
  publisher={Nature Publishing Group UK London}
}

@article{gProfiler,
  title={g: Profiler—a web-based toolset for functional profiling of gene lists from large-scale experiments},
  author={Reimand, J{\"u}ri and Kull, Meelis and Peterson, Hedi and Hansen, Jaanus and Vilo, Jaak},
  journal={Nucleic acids research},
  volume={35},
  number={suppl\_2},
  pages={W193--W200},
  year={2007},
  publisher={Oxford University Press}
}

@article{deepdtf,
  title={DeepDTF: Dual-Branch Transformer Fusion for Multi-Omics Anticancer Drug Response Prediction},
  author={Zhao, Yuhan and Tennant, Jacob and Yang, James and Guo, Zhishan and Whang, Young and Sui, Ning},
  journal={arXiv preprint arXiv:2603.24265},
  year={2026}
}

@article{madrigal,
  title={Multimodal AI predicts clinical outcomes of drug combinations from preclinical data},
  author={Huang, Yepeng and Su, Xiaorui and Ullanat, Varun and Moon, Intae and Liang, Ivy and Clegg, Lindsay and Olabode, Damilola and Johnson, Ruthie and Ho, Nicholas and Gibbs, Megan and others},
  journal={arXiv preprint arXiv:2503.02781},
  year={2025}
}

@article {precious3gpt,
	author = {Galkin, Fedor and Naumov, Vladimir and Pushkov, Stefan and Sidorenko, Denis and Urban, Anatoly and Zagirova, Diana and Alawi, Khadija M. and Aliper, Alex and Gumerov, Ruslan and Kalashnikov, Aleksandr and Mukba, Sabina and Pogorelskaya, Aleksandra and Ren, Feng and Shneyderman, Anastasia and Tang, Qiuqiong and Xiao, Deyong and Tyshkovskiy, Alexander and Ying, Kejun and Gladyshev, Vadim N. and Zhavoronkov, Alex},
	title = {Precious3GPT: Multimodal Multi-Species Multi-Omics Multi-Tissue Transformer for Aging Research and Drug Discovery},
	elocation-id = {2024.07.25.605062},
	year = {2024},
	doi = {10.1101/2024.07.25.605062},
	publisher = {Cold Spring Harbor Laboratory},
	URL = {https://www.biorxiv.org/content/early/2024/07/25/2024.07.25.605062},
	eprint = {https://www.biorxiv.org/content/early/2024/07/25/2024.07.25.605062.full.pdf},
	journal = {bioRxiv}
}

@article{scgpt,
  title={scGPT: toward building a foundation model for single-cell multi-omics using generative AI},
  author={Cui, Haotian and Wang, Chloe and Maan, Hassaan and Pang, Kuan and Luo, Fengning and Duan, Nan and Wang, Bo},
  journal={Nature methods},
  volume={21},
  number={8},
  pages={1470--1480},
  year={2024},
  publisher={Nature Publishing Group US New York}
}

@article{massgenie,
  title={Structural annotation of unknown molecules in a miniaturized mass spectrometer based on a transformer enabled fragment tree method},
  author={Yang, Yiming and Sun, Shuang and Yang, Shuyuan and Yang, Qin and Lu, Xinqiong and Wang, Xiaohao and Yu, Quan and Huo, Xinming and Qian, Xiang},
  journal={Communications Chemistry},
  volume={7},
  number={1},
  pages={109},
  year={2024},
  publisher={Nature Publishing Group UK London}
}

@article{mot,
  title={MOT: a Multi-Omics Transformer for multiclass classification tumour types predictions},
  author={Osseni, Mazid Abiodoun and Tossou, Prudencio and Laviolette, Francois and Corbeil, Jacques},
  journal={BioRxiv},
  pages={2022--11},
  year={2022},
  publisher={Cold Spring Harbor Laboratory}
}

@article{dnabert,
  title={DNABERT: pre-trained Bidirectional Encoder Representations from Transformers model for DNA-language in genome},
  author={Ji, Yanrong and Zhou, Zhihan and Liu, Han and Davuluri, Ramana V},
  journal={Bioinformatics},
  volume={37},
  number={15},
  pages={2112--2120},
  year={2021},
  publisher={Oxford University Press}
}

@article{nucleotidetransformer,
  title={Nucleotide transformer: building and evaluating robust foundation models for human genomics},
  author={Dalla-Torre, Hugo and Gonzalez, Liam and Mendoza-Revilla, Javier and Lopez Carranza, Nicolas and Grzywaczewski, Adam Henryk and Oteri, Francesco and Dallago, Christian and Trop, Evan and De Almeida, Bernardo P and Sirelkhatim, Hassan and others},
  journal={Nature Methods},
  volume={22},
  number={2},
  pages={287--297},
  year={2025},
  publisher={Nature Publishing Group US New York}
}

@article{data_kidney,
  title={Comprehensive proteogenomic characterization of rare kidney tumors},
  author={Li, Ginny Xiaohe and Chen, Lijun and Hsiao, Yi and Mannan, Rahul and Zhang, Yuping and Luo, Jie and Petralia, Francesca and Cho, Hanbyul and Hosseini, Noshad and da Veiga Leprevost, Felipe and others},
  journal={Cell Reports Medicine},
  volume={5},
  number={5},
  year={2024},
  publisher={Elsevier}
}

@article{data_glioblast,
  title={Proteogenomic and metabolomic characterization of human glioblastoma},
  author={Wang, Liang-Bo and Karpova, Alla and Gritsenko, Marina A and Kyle, Jennifer E and Cao, Song and Li, Yize and Rykunov, Dmitry and Colaprico, Antonio and Rothstein, Joseph H and Hong, Runyu and others},
  journal={Cancer cell},
  volume={39},
  number={4},
  pages={509--528},
  year={2021},
  publisher={Elsevier}
}
\bibliographystyle{icml2026}

\newpage
\appendix
\onecolumn
\section*{Appendix}

\section{Model Architecture} \label{architecturedetails}
The architecture of ARO is shown in Figure \ref{fig:architecture}. The model is implemented as a fully-connected feedforward autoencoder. The encoder consists of an input dropout layer ($p=0.1$), followed by a linear projection from 71,094 to 2,048 dimensions and three fully connected hidden layers of dimension 2,048. Each hidden layer is followed by a ReLU activation and dropout ($p=0.0$). The result is a 2,048-dimensional hidden representation of modalities. There is a reduction in the number of features in the model due to dropping of zero-variance features during processing, also outlined in Sec \ref{sec:data_process}.

The decoder reconstructs the original feature space through a final linear layer projection from the latent representation back to the original dimensions, followed by a softplus activation to enforce non-negative outputs after preprocessing and log-space normalization.

The training pipeline is implemented using PyTorch Lightning, with Hydra for configuration management and Weights \& Biases for experiment tracking and hyperparameter tuning.
The model evaluated in Sec. \ref{sec:results} has a total of $\mathtt{\sim}$303.9M trainable parameters and the outline is reported in Table \ref{paramstable}.

\begin{table}[h]
  \caption{Parameter numbers of each block of the autoencoder.}
  \label{paramstable}
  \begin{center}
        \begin{sc}
            \begin{tabular}{lcc}
              \toprule
                Layer & Shape & Parameters  \\
              \midrule
              inp\_lin.weight  & (2048, 71094) & 145,600,512  \\
              inp\_lin.bias   & (2048, ) & 2048  \\
              encoder\_layers.0.weight   & (2048, 2048) & 4,194,304  \\
              encoder\_layers.0.bias   & (2048, ) & 2048 \\
              encoder\_layers.1.weight   & (2048, 2048) & 4,194,304  \\
              encoder\_layers.1.bias   & (2048, ) & 2048 \\        
              encoder\_layers.2.weight   & (2048, 2048) & 4,194,304  \\
              encoder\_layers.2.bias   & (2048, ) & 2048 \\
              out\_lin.weight   & (71094, 2048) & 145,600,512  \\
              out\_lin.bias   & (71094, ) & 71094 \\
              \bottomrule
            \end{tabular}
        \end{sc}
  \end{center}
  \vskip -0.1in
\end{table}

\section{Optimization Details and Hyperparameter Tuning} \label{hyperparamtuning}
Optimization is performed using AdamW with a learning rate of $10^{-5}$ and no weight decay. A ReduceLROnPlateau scheduler is employed with a reduction factor of 0.1 and patience of 10 epochs. Training is performed for 200 epochs with a batch size of 80 using mixed bfloat16 precision (fp32 as fallback on unsupported hardware) and JIT compilation via torch.compile on CUDA-enabled devices. No batch normalization is employed.

\begin{figure}[h]
    \centering
    \begin{subfigure}{\linewidth}
        \includegraphics[width=\linewidth]{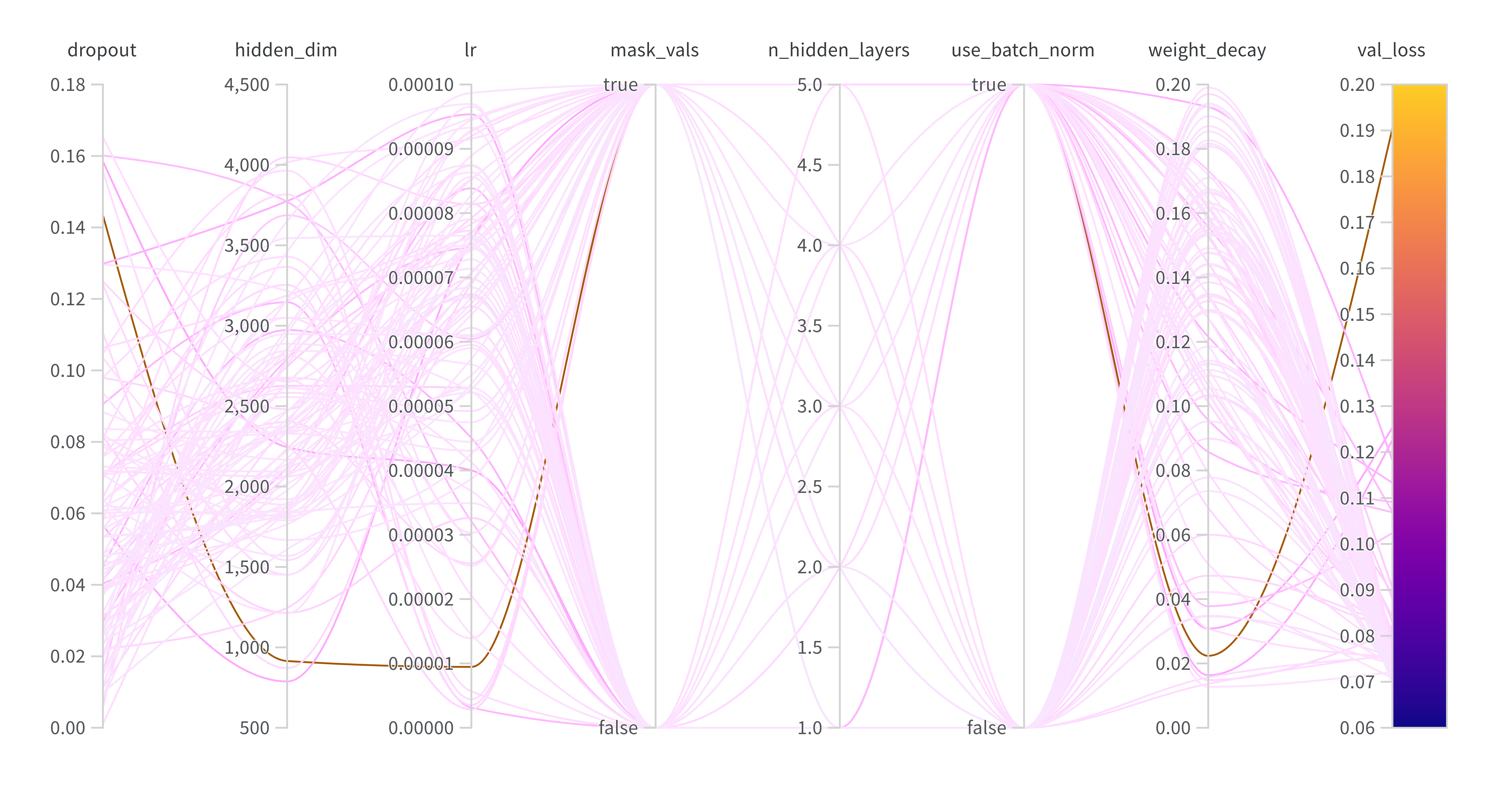}
        \caption{Using dropout on input}
    \end{subfigure}
    \begin{subfigure}{\linewidth}
        \includegraphics[width=\linewidth]{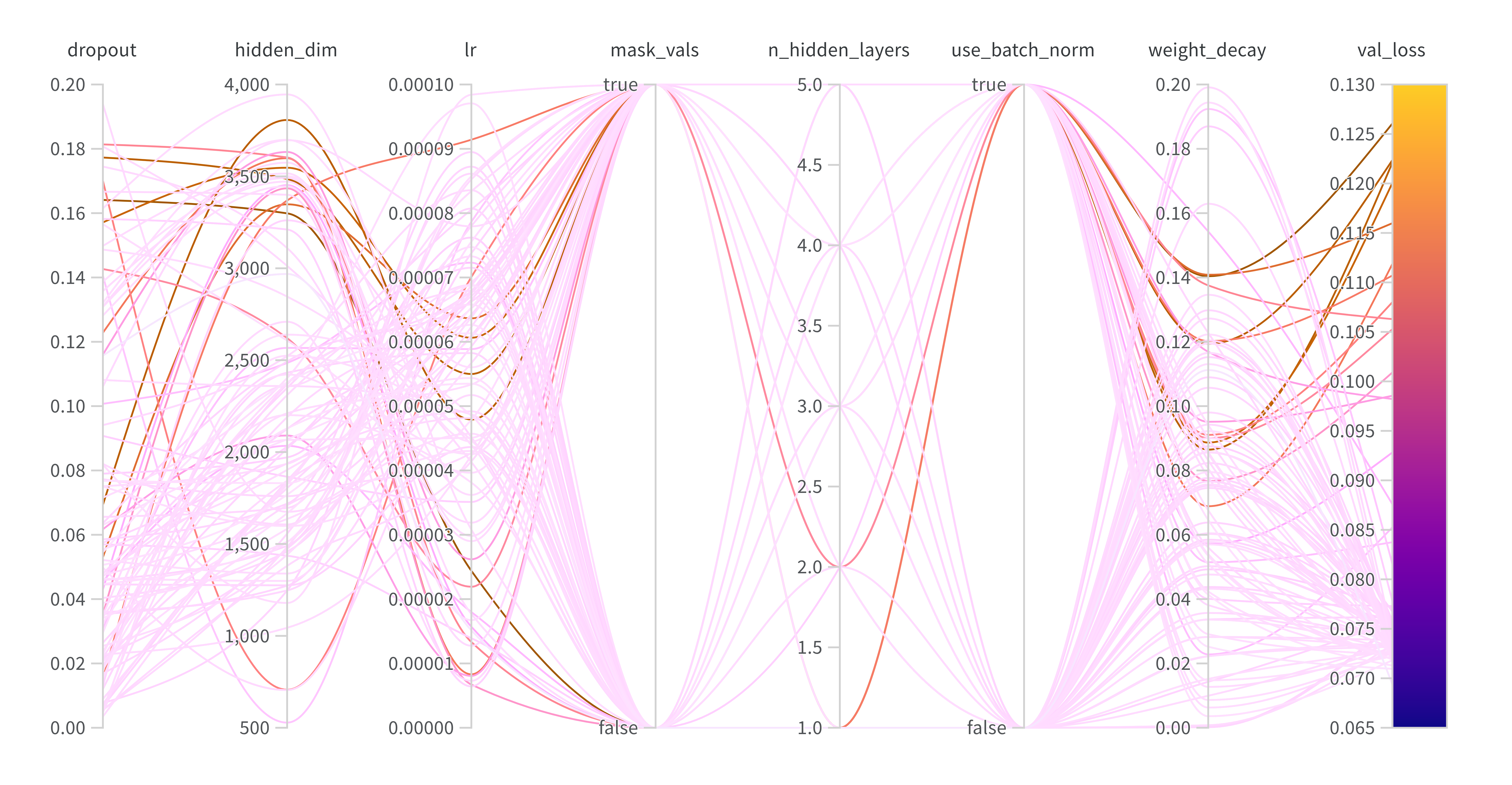}
        \caption{Not using dropout on input}
    \end{subfigure}
    \caption{Iterations for hyperparameter tuning performed over validation loss, optimizing for dropout configuration, hidden dimensions, learning rate, modality-specific masking, number of hidden layers, batch normalization usage and weight decay. The configuration with the lowest validation loss is chosen from each setup.}
    \label{fig:hyperparamtuning}
\end{figure}

Hyperparameter tuning was performed via Bayesian Optimization through Weights \& Biases. The parameters optimized for are: dropout configuration, hidden dimensions, learning rate, modality-specific masking, number of hidden layers, batch normalization usage and weight decay---and the results are shown in Figure \ref{fig:hyperparamtuning}. 

During training, the model optimizes for the Huber loss, defined as:
\begin{equation}
L_\delta(y, \hat{y}) =
\begin{cases}
\frac{1}{2}(y - \hat{y})^2 & \text{if } |y - \hat{y}| \le \delta \\
\delta \left( |y - \hat{y}| - \frac{1}{2}\delta \right) & \text{otherwise}
\end{cases} \label{eq:huberloss}
\end{equation}

\section{Autoencoders}
The model exploits the Autoencoder (AE) architecture \cite{autoencoders}. AEs are unsupervised neural networks designed to learn efficient data encodings by forcing the model to reconstruct the input from a compressed latent representation (i.e. the ``bottleneck"). An autoencoder consists of both an encoder and a decoder network, whereby the encoder first compresses the input $x \in \mathbb{R}^n$ into a latent space representation $h \in \mathbb{R}^m$ . The decoder then maps this latent representation back to the original input space to produce a reconstruction $y \in \mathbb{R}^n$:
\begin{equation}
    h = f_e(x) = s_e(W_e x + b_e)
\end{equation}
\begin{equation}
    y = f_d(h) = s_d(W_d h + b_d)
\end{equation} 
where $s_{\{e,d\}}$ are the encoder/decoder activation functions, $W_{\{e,d\}} \in \mathbb{R}^{m \times n}$ are the encoder/decoder weight matrices, and $b_{\{e,d\}} \in \mathbb{R}^m$ are the encoder/decoder biases.

The objective of the AE is to minimize the reconstruction loss $L_\delta$ (Huber loss in our case, defined in Eq. \ref{eq:huberloss}) over a dataset of $N$ samples, expressed as:
\begin{equation}
    \argmin J\left(W_{\{e,d\}}, b_{\{e,d\}}\right) = \frac{1}{N} \sum_{i=1}^{N} L_\delta\left(x^{(i)}, y^{(i)}\right)
\end{equation}

\section{Results}
\begin{table}[h]
  \caption{Reconstruction performance using MSE Loss averaged over 5 runs}
  \label{tab:appndx_mse_comparison}
  \centering
      \begin{sc}
        \begin{tabular}{lccc}
          \toprule
             & Train & Validation & Test \\
          \midrule
          ARO Masked  & $0.046 \pm 56.0e-04$ & $0.14 \pm 31.0e-05$ & $0.14 \pm 21.0e-05$ \\
          ARO Unmasked  & $0.064 \pm 31.0e-05 $ & $0.15 \pm 21.0e-05$ & $0.15 \pm 24.0e-05$ \\
          PCA  & $3.1e-30$ & 0.14 & 0.14 \\
          \bottomrule
        \end{tabular}
    \end{sc}
  \vskip -0.1in
\end{table}

\begin{table}[h]
  \caption{Reconstruction performance using Huber Loss averaged over 5 runs}
  \label{tab:appndx_huber_comparison}
  \centering
        \begin{sc}
        \begin{tabular}{lccc}
          \toprule
             & Train & Validation & Test \\
          \midrule
          ARO Masked  & $0.022 \pm 28.0e-04$ & $0.070 \pm 5.6e-05$ & $0.068 \pm 20.0e-05$ \\
          ARO Unmasked  & $0.032 \pm 16.0e-05$ & $0.071 \pm 10.0e-05 $ & $0.07 \pm 6.1e-05$ \\
          PCA  & $1.5e-30$ & 0.068 & 0.065 \\
          \bottomrule
        \end{tabular}
        \end{sc}
  \vskip -0.1in
\end{table}

\begin{figure}[!h]
    \centering
    \includegraphics[width=0.5\linewidth]{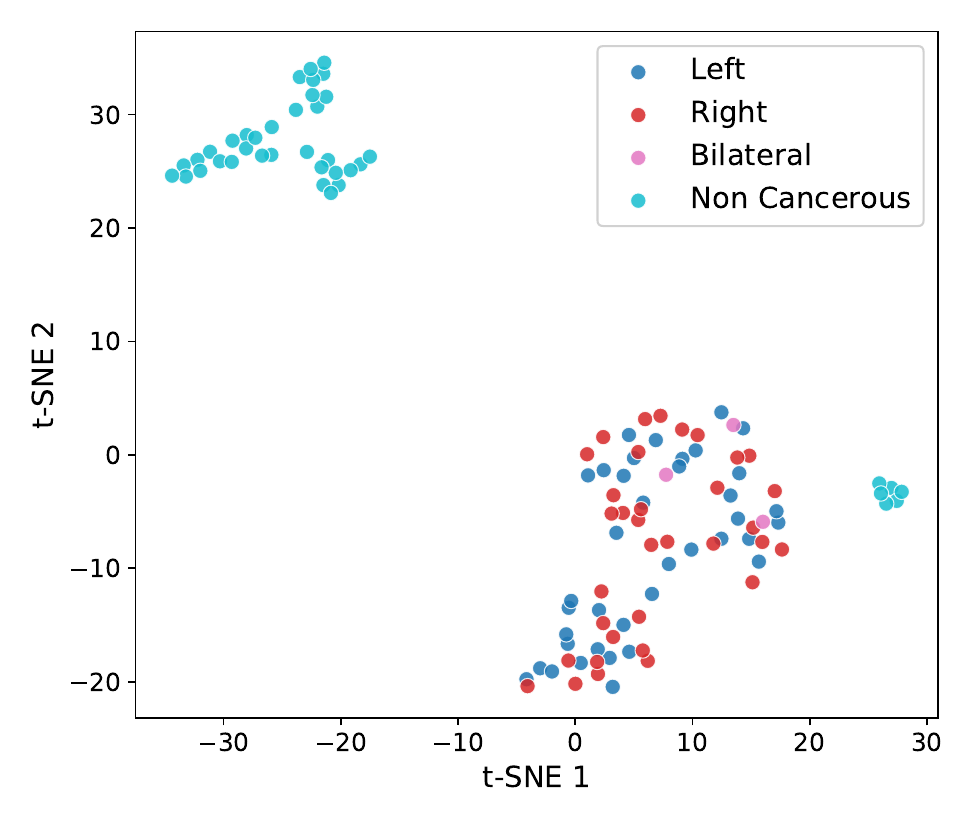}
    \caption{t-SNE cluster of the learned embeddings from ARO}
    \label{fig:tsne_cluster}
\end{figure}

\section{Comparison}
\begin{figure}[h]
    \centering
    \begin{subfigure}{.45\linewidth}
        \includegraphics[width=\linewidth]{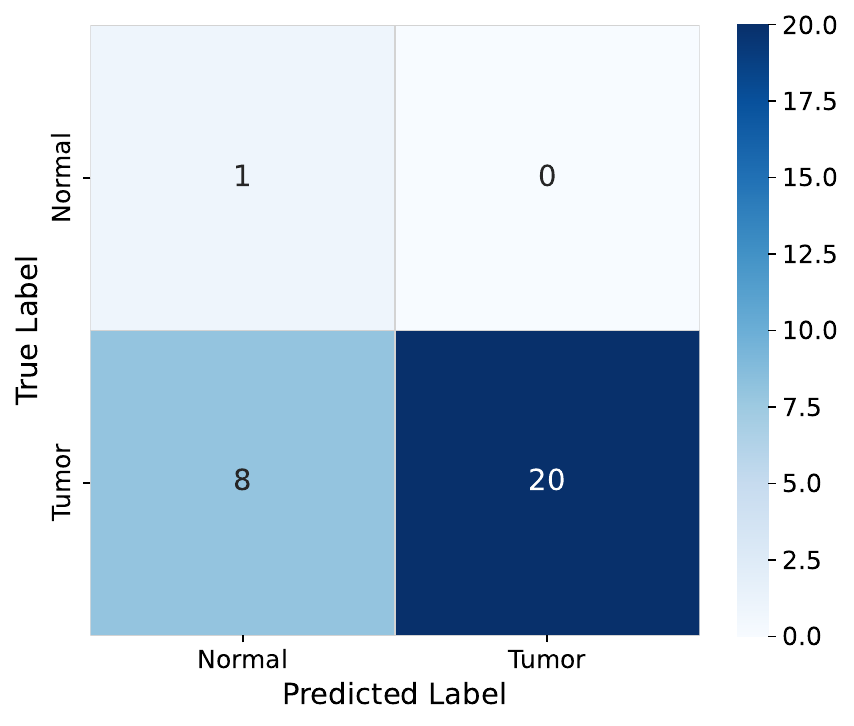}
        \caption{Binary classification task}
        \label{fig:pca_binary_conf_mat}
    \end{subfigure}
    \begin{subfigure}{.45\linewidth}
        \includegraphics[width=\linewidth]{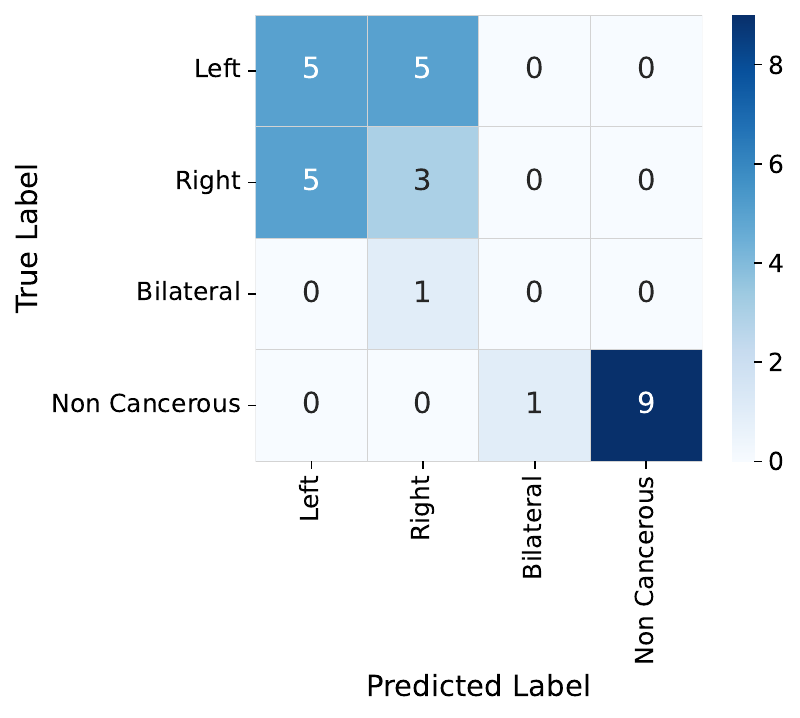}
        \caption{Multi-label classification task}
        \label{fig:pca_multi_conf_mat}
    \end{subfigure}
    \caption{Confusion matrices of downstream tasks using PCA projections.}
    \label{fig:pca_comparison_conf_matrix}
\end{figure}

Confusion matrix of the downstream analysis using PCA is given in Figure \ref{fig:pca_comparison_conf_matrix}.

\section{Dataset Breakdown}

\begin{table}[h]
  \caption{\textbf{Details of the Multi-omics} datasets considered and analysed in the study.}
  \label{datasets}
  \begin{center}
      \begin{sc}
        \begin{tabular}{lcccr}
          \toprule
          Data set & Samples & Feature Count  \\
          \midrule
          \cite{data_kidney} - Transcriptomics      & 108 & 45886 \\
          \cite{data_kidney} - Proteomics           & 109 & 10998  \\
          \cite{data_kidney} - Metabolomics         & 87 & 67  \\
          \cite{data_glioblast} - Transcriptomics   & 240 & 26941 \\
          \cite{data_glioblast} - Proteomics        & 333 & 12309 \\
          \cite{data_glioblast} - Metabolomics      & 36 & 216 \\
          \bottomrule
        \end{tabular}
      \end{sc}
  \end{center}
  \vskip -0.1in
\end{table}

\clearpage

\end{document}